\documentclass[10pt,twocolumn,letterpaper]{article}

\usepackage{iccv}
\usepackage{times}
\usepackage{epsfig}
\usepackage{graphicx}
\usepackage{amsmath}
\usepackage{amssymb}
\usepackage{multirow}
\usepackage{algorithm,epstopdf,algorithmic,subfigure,booktabs,ctable}
\usepackage{enumerate}

\usepackage[breaklinks=true,bookmarks=false]{hyperref}

\iccvfinalcopy % *** Uncomment this line for the final submission

\def\iccvPaperID{****} % *** Enter the ICCV Paper ID here

\begin{document}
	
	%%%%%%%%% TITLE
	\title{Multi-Granularity Fusion Network for Proposal and Activity Localization: Submission to ActivityNet Challenge 2019 Task 1 and Task 2}
	
	\author{Haisheng Su, Shuming Liu, Xu Zhao\thanks{Corresponding author.}  \\
		\\
		Department of Automation,\\
		Shanghai Jiao Tong University\\
		{\tt\small \{suhaisheng,shumingliu,zhaoxu\}@sjtu.edu.cn}
		\\
		% For a paper whose authors are all at the same institution,
		% omit the following lines up until the closing ``}''.
		% Additional authors and addresses can be added with ``\and'',
		% just like the second author.
		% To save space, use either the email address or home page, not both
	}
	
	\maketitle
	%\thispagestyle{empty}

	%%%%%%%%% ABSTRACT
	\begin{abstract}
		This technical report presents an overview of our solution used in the submission to ActivityNet Challenge 2019 Task 1 (\textbf{temporal action proposal generation}) and Task 2 (\textbf{temporal action localization/detection}). Temporal action proposal indicates the temporal intervals containing the actions and plays an important role in temporal action localization. Top-down and bottom-up methods are the two main categories used for proposal generation in the existing literature. In this paper, we devise a novel Multi-Granularity Fusion Network (MGFN) to combine the proposals generated from different frameworks for complementary filtering and confidence re-ranking. Specifically, we consider the diversity comprehensively from multiple perspectives, e.g. the characteristic aspect, the data aspect, the model aspect and the result aspect. Our MGFN achieves the state-of-the-art performance on the temporal action proposal task with 69.85 AUC score and the temporal action localization task with 38.90 mAP on the challenge testing set.
	\end{abstract}
	
	%%%%%%%%% BODY TEXT
	\section{Task Introduction}
	Temporal action detection task has received much attention from many researchers in recent years, which requires not only categorizing the real-world untrimmed videos but also locating the temporal boundaries of action instances. Analogous to object proposals for object detection in images, temporal action proposal indicates the temporal intervals containing the actions and plays an important role in video temporal action detection. 
	It has been commonly recognized that high-quality proposals should have precise temporal boundaries and reliable confidence scores. To cater for these two conditions and achieve high quality proposals, there are two main categories in the existing proposal generation methods \cite{sst_buch_cvpr17,escorcia2016daps,gao2017turn,SSAD,SCNN}. 
	However, the proposals generated in a \textit{top-down} fashion are doomed to have imprecise boundaries though with regression. Under this circumstance, the other type of methods \cite{BSN,Y.Xiong} have drawn much attention in the community recently which tackle this problem in a \textit{bottom-up} fashion, where the input video is evaluated in a finer-level. \cite{BSN} is a typical method in this type which proposes the Boundary Sensitive Network (BSN) to generate proposals with flexible durations and reliable confidence scores. Though BSN achieves convincing performance in this manner, it still suffers from many drawbacks. For example, the snippet-level probability sequence of actionness or boundary is sensitive to noises and the inferior quality of confidence score used for proposal retrieving. 
	
	\section{Approach Overview}
	In this section, we will introduce the technical details of our approach.
	%-------------------------------------------------------------------------
	\subsection{Video Features Encoding}
	we adopt the two-stream network \cite{K.Simonyan} in advance to encode the visual features of an input video, where the RGB stream handles a RGB image as input to capture the spatial features, while the flow stream operates on the stacked optical flows to capture the motion information. This kind of architecture has been widely used in action recognition \cite{TSN} and temporal action detection tasks. As for the characteristic aspect, we try many different ConvNet architectures pre-trained on Kinetics-400 dataset, such as ResNet-50, ResNet-101, ResNet-152, ResNet-200, I3D, P3D, Inception-V3 and Inception-ResNet-V2, to verify the effectiveness, which are then used for feature extraction. Finally, we employ a set of effective feature representations for proposal generation, thus to ensure the feature diversity. 
	
	\subsection{Data Augmentation}
	Considering the ground-truth distribution and in order to reduce computational cost, we rescale the length of each feature sequence to a fixed size by linear interpolation before feeding it into our MGFN. As for data augmentation and varied lengths of action instances, we adopt a set of lengths of feature sequence during training phase, such as 64, 100, 128 and 192. Meanwhile, we randomly sample 2000 validation videos and add them to the training set, while leave others for validation.
	
	\subsection{Multi-Granularity Fusion Network}
	
	\textbf{APN}. Prop-SSAD \cite{TCAP} is a simplified version of SSAD \cite{SSAD} and is the first to perform anchor mechanism on the temporal action proposal generation task, which utilizes several temporal convolution anchor layers with different resolutions to generate proposals with varied lengths. The lower anchor layers are used to locate the short-range action proposals while the higher anchor layers are responsible to cover the long-range action proposals. Through this mechanism, the generated proposals can be densely distributed on each feature map. In this paper, we improve the rank performance with three types of classifiers, namely a binary activity classifier, a completeness classifier and an Intersection-over-Union (\textit{IoU}) classifier/regressor. 
	
	In selecting positive samples for activity classifier during training, proposals that overlap with a ground-truth instance with an \textit{IoU} larger than 0.7 or lower than 0.3 but the intersection with a ground-truth over its own time span (\textit{IoP}) above 0.8 will be used. While the proposals are regarded as negative samples only when less than 5\% of its time span overlaps with any ground-truth action instances. 
	
	As for the completeness classifier, proposals with \textit{IoU} larger than 0.7 are employed as positive samples, and proposals with $ IoU < 0.3 $ while $ IoP > 0.8 $ are used as negative samples. As for the \textit{IoU} classifier, we divide the proposals into three categories according to the \textit{IoU} values. The value range 0-1 is discretized into three ranges $ \{$0-0.3, 0.3-0.7, 0.7-1.0$\}$, referred as the background value range, the middle value range and the high value range respectively. With these three classifiers, our Anchor Pyramid Network (APN) can evaluate the proposals comprehensively with complicated situations, and we use 7 anchor layers to predict the proposals with 512 feature maps. During inference stage, we fuse the outputs of three classifiers to obtain the confidence score $ p_{conf} $ for each proposal:
	\begin{equation}
	p_{conf} = p_{a}\cdot p_{c}\cdot p_{i},
	\end{equation}
	where $ p_{a}, p_{c}, p_{i} $ indicate the actionness score, the completeness score and the \textit{IoU} score respectively. However, the proposals generated in this way are doomed to have imprecise boundaries though with regression. 
	
	\textbf{TAG}. TAG \cite{Y.Xiong} first evaluates the snippet-level actionness indicating whether the snippet is inside the action instances, then adopts watershed algorithm to group the consecutive snippets with two set of thresholds. Proposals generated in this \textit{bottom-up} fashion are more sensitive to the temporal boundaries than anchor-based methods. However, proposals generated by TAG can not be further retrieved without confidence scores evaluated in a global view.
	
	\textbf{Improved-BSN}. BSN \cite{BSN} also generates the proposals in a \textit{bottom-up} fashion which first evaluates the probabilities of each temporal location being in the starting, ending and middling regions. Then through combining the high probability boundary locations, it can generate abundant proposals with flexible durations and confidence scores. However, the probability sequences predicted by a simple three-layer temporal convolution network are sensitive to noises, causing many false alarms and low precision. Besides, the performance of confidence scores used for proposal retrieving are also limited owing to the inferior proposal-level representations. In this paper, we further promote the performance of BSN through improving the quality of probability sequence with several edge-smoothing strategies and the proposal-level representations used for ranking. Besides, we unify the training process for a robust optimization.

	\textbf{Complementary Filtering, Temporal Boundary Adjustment and Proposal Ranking Model (CAR)}. As we discussed above, proposals generated by the actionness score grouping method and the anchor/sliding window based method are complementary with each other. Specifically, the proposals generated by the anchor-based method can uniformly cover the whole videos while with imprecise boundaries, and the grouping based method can generate proposals with more precise boundaries but rely greatly on the actionness score, especially when the actionness score is low, it may dismiss some potential proposals. Under this circumstance, we first train a binary classifier with ground-truth action instances as input, while use the proposal results of TAG as label set. In selecting the positive samples, if the ground-truth instances overlap with a proposal of TAG with an \textit{IoU} larger than 0.5, the input ground-truths will be labeled as 1, otherwise 0. During testing phase, we feed the proposal results of APN to the binary classifier in order to select the proposals with low scores ($ < $ 0.5). Then the selected proposals from APN are combined with TAG for boundary adjustment and proposal ranking, with three-stage (left, central, right) unit features as input to the multi-layer perceptron model respectively.

	\textbf{Proposal Re-ranking}. Since both the proposal generation and quality evaluation can influence the evaluation of proposals, we re-rank the proposals of improved BSN as:
	\begin{equation}
	p_{conf}^{\tau} = p_{s}^{\tau}\cdot p_{e}^{\tau}\cdot p_{iou}^{\tau}\cdot p_{match}^{\tau},
	\end{equation}
	where $ p_{s}^{\tau}, p_{e}^{\tau}, p_{iou}^{\tau} $ indicate the starting probability, ending probability and \textit{IoU} score of the proposal $ \tau $ predicted by the improved-BSN. And $ p_{match}^{\tau} $ indicates the confidence of the proposal of APN which has the maximum $ IoU $ with $ \tau $.

	%-------------------------------------------------------------------------
	\section{Experiment Results}
	\subsection{Evaluation Metrics}
	For temporal action proposal generation task, Average Recall (AR) calculated under different \textit{tIoU} thresholds is commonly adopted as one evaluation metric. In this challenge, the thresholds are set from 0.5 to 0.95 with a step size of 0.05. And the area under the Average Recall vs. Average Number of Proposals (AN) curve (AUC) are used as the final evaluation metric, where AN ranges from 0 to 100.
	
	For temporal action localization task, mean Average Precision (mAP) is a conventional evaluation metric, where Average Precision (AP) is calculated for each category respectively. In this challenge, the average mAP with \textit{tIoU} thresholds from 0.5 to 0.95 with a step size of 0.05 is reported.

	\subsection{Temporal Action Proposal and Localization}
	The performance of APN, TAG, improved-BSN and CAR on the validation set of ActivityNet-1.3 are shown in the Table 1. For model fusion process, we consider the inter-model and intra-model fusion respectively. We obtain the results of each model with multi-modality fusion of different feature representations and testing scales. As for multi-scale testing, we not only concatenate the outputs of different fixed scales  but also the free scales of videos. As shown in Fig. 1, we illustrate the difference of ground-truth AUC distribution between fix-scale and free-scale testing on the validation set of ActivityNet-1.3. And we can observe that the ground-truth AUC contribution of free-scale testing is superior than fix-scale testing when the duration of ground-truth is relatively short. What's more, we further perform the score fusion between the improved-BSN and APN to re-rank the proposals. 
	
	For task 1, through merging the improved-BSN \& APN model and CAR model by Soft-NMS \cite{Bodla2017Improving}, we achieve 69.85 AUC score on both validation set and testing server. 
	
	For task 2, our improved-BSN \& APN achieves 38.90 mAP on the testing server and win the \textbf{third place} of temporal action localization task in ActivityNet Challenge 2019.

	\begin{figure}[t]
		\begin{center} 
			\includegraphics[width=1.1\linewidth]{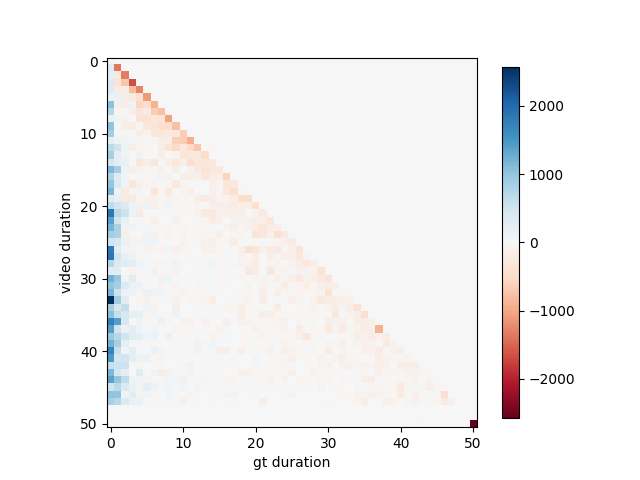}
		\end{center}
		\caption{Visualization of the difference of ground-truth AUC distribution between fix-scale and free-scale testing on the validation set of ActivityNet-1.3 dataset. Note that all scale values along the axis should be multiplied by 5 to represent the second.}
		\label{fig:long-term}
	 
	\end{figure}

	\section{Conclusion}
	In this challenge technical notebook, we comprehensively analyze the complementary characteristics of \textit{bottom-up} and \textit{top-down} proposal generation methods, and our enhanced APN, BSN and CSR models all contribute to the performance improvement. Specifically, with the smoothed probability sequence, our improved-BSN can generate the proposals with higher precision. And with three additional classifiers, our improved-APN can evaluate the proposals more reasonably. All these improvements can also reveal the direction of how to make better temporal action proposal generation and localization.

		\setlength{\tabcolsep}{6pt}
	\begin{table}
		\begin{center}
			\caption{Proposal results on validation set of ActivityNet-1.3.}
			\begin{tabular}{p{4cm}|c}
				\toprule
				Setting & AUC (val) \tabularnewline
				\noalign{\smallskip}
				\hline  
				\noalign{\smallskip}
				
				APN & 62.45  \tabularnewline
				TAG & 63.97  \tabularnewline
				improved-BSN  & 68.18  \tabularnewline
				imporved-BSN+APN  & 68.58  \tabularnewline
				CAR & 68.01   \tabularnewline
				improved-BSN+APN+CAR & \textbf{69.85}  \tabularnewline
				\bottomrule 
			\end{tabular}
			\label{table_comparison_features}
		\end{center}
		%\vspace{-0.4cm}
	\end{table}

	%-------------------------------------------------------------------------

	{\small
		\bibliographystyle{ieee}
		\bibliography{anet19_report}

\begin{thebibliography}{10}\itemsep=-1pt

\bibitem{Bodla2017Improving}
N.~Bodla, B.~Singh, R.~Chellappa, and L.~S. Davis.
\newblock Improving object detection with one line of code.
\newblock {\em arXiv preprint arXiv:1704.04503}, 2017.

\bibitem{sst_buch_cvpr17}
S.~Buch, V.~Escorcia, C.~Shen, B.~Ghanem, and J.~C. Niebles.
\newblock Sst: Single-stream temporal action proposals.
\newblock In {\em 2017 IEEE Conference on Computer Vision and Pattern
  Recognition (CVPR)}, pages 6373--6382. IEEE, 2017.

\bibitem{escorcia2016daps}
V.~Escorcia, F.~C. Heilbron, J.~C. Niebles, and B.~Ghanem.
\newblock Daps: Deep action proposals for action understanding.
\newblock In {\em European Conference on Computer Vision}, pages 768--784.
  Springer, 2016.

\bibitem{gao2017turn}
J.~Gao, Z.~Yang, C.~Sun, K.~Chen, and R.~Nevatia.
\newblock Turn tap: Temporal unit regression network for temporal action
  proposals.
\newblock In {\em Computer Vision (ICCV), 2017 IEEE International Conference
  on}, pages 3648--3656. IEEE, 2017.

\bibitem{SSAD}
T.~Lin, X.~Zhao, and Z.~Shou.
\newblock Single shot temporal action detection.
\newblock In {\em Proceedings of the 2017 ACM on Multimedia Conference}, pages
  988--996. ACM, 2017.

\bibitem{TCAP}
T.~Lin, X.~Zhao, and Z.~Shou.
\newblock Temporal convolution based action proposal: Submission to activitynet
  2017.
\newblock {\em arXiv preprint arXiv:1707.06750}, 2017.

\bibitem{BSN}
T.~Lin, X.~Zhao, H.~Su, C.~Wang, and M.~Yang.
\newblock Bsn: Boundary sensitive network for temporal action proposal
  generation.
\newblock {\em arXiv preprint arXiv:1806.02964}, 2018.

\bibitem{SCNN}
Z.~Shou, D.~Wang, and S.-F. Chang.
\newblock Temporal action localization in untrimmed videos via multi-stage
  cnns.
\newblock In {\em CVPR}, pages 1049--1058, 2016.

\bibitem{K.Simonyan}
K.~Simonyan and A.~Zisserman.
\newblock Two-stream convolutional networks for action recognition in videos.
\newblock In {\em Advances in neural information processing systems}, pages
  568--576, 2014.

\bibitem{TSN}
L.~Wang, Y.~Xiong, Z.~Wang, Y.~Qiao, D.~Lin, X.~Tang, and L.~Van~Gool.
\newblock Temporal segment networks: Towards good practices for deep action
  recognition.
\newblock In {\em ECCV}, pages 20--36. Springer, 2016.

\bibitem{Y.Xiong}
Y.~Xiong, Y.~Zhao, L.~Wang, D.~Lin, and X.~Tang.
\newblock A pursuit of temporal accuracy in general activity detection.
\newblock {\em arXiv preprint arXiv:1703.02716}, 2017.

\end{thebibliography}
	}
	
\end{document}